\documentclass[runningheads]{sty/llncs}
\usepackage{graphicx}
\pdfoutput=1 
\usepackage{tikz}
\usepackage{comment}
\usepackage{amsmath,amssymb} 
\usepackage{color}
\usepackage{wrapfig}
\usepackage{multirow}
\usepackage{booktabs}
\usepackage{hyperref}
\hypersetup{colorlinks,allcolors=black}

\usepackage{amssymb}
\usepackage{pifont}

\begin{document}
\pagestyle{headings}
\mainmatter
\def\ECCVSubNumber{6236}  

\title{SWFormer: Sparse Window Transformer for 3D Object Detection in Point Clouds} 

\titlerunning{SWFormer}

\author{Pei Sun \and
Mingxing Tan \and
Weiyue Wang \and
Chenxi Liu \and
Fei Xia \and
Zhaoqi Leng \and
Dragomir Anguelov}
\authorrunning{P. Sun et al.}

\institute{
Waymo LLC \\
\email{\{peis, tanmingxing, weiyuewang, cxliu, feixia, lengzhaoqi, dragomir\}\\@waymo.com}
}
\maketitle

\newcommand{\ourmethod}{SWFormer}
\newcommand{\cmark}{\ding{51}}%
\newcommand{\xmark}{\ding{55}}%
\newcommand{\blue}[1]{\color{blue} {#1}}
\newcommand{\cyan}[1]{\color{cyan} {#1}}

\begin{abstract}

3D object detection in point clouds is a core component for modern robotics and autonomous driving systems. A key challenge in 3D object detection comes from the inherent \emph{sparse} nature of point occupancy within the 3D scene. In this paper, we propose Sparse Window Transformer (\emph{\ourmethod}), a scalable and accurate model for 3D object detection, which can take full advantage of the sparsity of point clouds. Built upon the idea of window-based Transformers, {\ourmethod} converts 3D points into sparse voxels and windows, and then processes these variable-length sparse windows efficiently using a bucketing scheme. In addition to self-attention within each spatial window, our {\ourmethod} also captures cross-window correlation with multi-scale feature fusion and window shifting operations. To further address the unique challenge of detecting 3D objects accurately from sparse features, we propose a new voxel diffusion technique. Experimental results on the Waymo Open Dataset show our {\ourmethod} achieves state-of-the-art \emph{73.36} L2 mAPH on vehicle and pedestrian for 3D object detection on the official test set, outperforming all previous single-stage and two-stage models, while being much more efficient.

  
\end{abstract}

\section{Introduction}

3D point cloud representation learning is critical for autonomous driving, especially for core tasks like 3D object detection. The challenges of learning from 3D point clouds mainly come from two aspects. The first aspect is that 3D points are sparsely distributed in the 3D space due to the nature of LiDAR sensors. This forces 3D models to be different from dense models in natural language processing (where words in a sentence are dense) or image understanding (where pixels in an image are dense). The second aspect is that both the number of points in a point cloud frame and the point cloud sensing region are increasing along with the improvement of the LiDAR sensor hardware. Some of the latest commercial LiDARs can sense up to 250m~\cite{packnet} and 300m~\cite{WaymoV5} in all directions around the vehicle, leading to a large range of point clouds.

To address these challenges, previous works have proposed many methods that can be roughly organized as five categories. 
\textbf{PointNet} \cite{qi2017pointnet,qi2017pointnet++,shi2019pointrcnn} based method treats 3D point clouds as unordered sets and encodes them with MLPs and max pooling. Hierarchical structure is introduced to deal with the large input space and to better capture local information. These methods usually have inferior representation capacity compared with more recent methods. \textbf{PointPillars}-style methods \cite{lang2019pointpillars} divide the space into grids of fixed sizes to convert the sparse 3D problem to a dense 2D problem. This method scales quadratically with the range, making it hard to scale with the advancement of LiDAR hardwares. \textbf{Sparse submanifold convolutions} \cite{graham2017submanifold,shi2020pv,sun2021rsn} based method can handle the sparse input efficiently. Usually these methods use small $3\times3$  convolution kernels which cannot connect features that are sparsely disconnected without adding normal sparse convolution and striding. This weakness limits its representation capacity. Another weakness of this method is their need for heavily optimized custom ops to be efficient on the modern GPUs and incompatibility with matmul optimized accelerators such as TPUs.  
\textbf{Range image} is a compact representation of point cloud. Multi-view methods \cite{zhou2019end,sun2021rsn,wang2020pillar,bewley2020range} run dense convolutions in this view to extract features and fuse with BEV features learned in the PointPillars-style to improve 3D representation learning. It is hard to regress 3D objects directly from the range image due to its lack of 3D information encoding in the dense 2D perspective convolutions. To tackle this weakness, graph-style kernels \cite{chai2021point,fan2021rangedet} replace convolutions to make use of the range information in range images to capture 3D information which greatly improves the accuracy but is still inferior to the state of the art. \textbf{Transformer} \cite{vaswani2017attention} is designed to process sequences of data. The challenge in applying it to a point cloud is to solve the quadratic complexity on the number of inputs. Recent methods tackle this problem by attending to neighboring points \cite{pan20213pointformer}, neighboring voxels \cite{mao2021voxel} or voxels in fixed windows \cite{singlestride21}. A generic and efficient transformer-only model without limitations like limited receptive field, irregular memory access pattern, and lack of scalability is still to be designed.

In this paper, we adapt window-based Transformers to 3D point clouds. The Transformer~\cite{vaswani2017attention} architecture has been hugely successful in modeling language sequences and image patches.
In particular, on 2D images, Swin Transformer~\cite{swin21} proposed to partition images into windows and merge context information in a hierarchical manner. Our \emph{Sparse Window Transformer ({\ourmethod})} builds upon similar ideas, but with several key adaptations for sparse windows. Our first adaption is to add a bucketing-based window partition for sparse windows. Although each window has the same spatial size, such as a $10\times10$ voxel grid, the number of non-empty voxels in each window can vary significantly, so we group these windows into buckets with different effective sequence lengths.  Our second adaptation is to limit the expensive window shifting. Swin Transformer~\cite{swin21} uses window shifting once per Transformer layer to connect features between windows and increases receptive fields, but this shifting operation is expensive in the sparse world as it needs to re-order all the sparse features with gather operations. Moreover, it is extremely slow on matmul optimized accelerators such as TPUs. To address this issue, {\ourmethod} employs a new hierarchical backbone architecture, where each {\ourmethod} block has many Transformer layers but only one shifting operation, as shown in \autoref{fig:swformerblock}. It relies on multi-scale features to achieve large receptive fields for context information, and a multi-scale fusion network to effectively combine these features. The model uses additional custom downsample and upsample algorithms to properly handle the sparse features during feature fusion.


Our innovation continues from the backbone into the 3D object detection head. Existing 3D object detection methods \cite{zhou2018voxelnet,lang2019pointpillars,zhou2019end,wang2020pillar,ge2020afdet,shi2020pv,chai2021point,sun2021rsn,mao2021voxeltransformer,yin2021centerpoint} can mostly be viewed as either anchor based methods with implicit or explicit anchors or DETR \cite{detr} based methods \cite{3ddetr}. The detection performance is closely related with the distribution of the difference between anchor and groundtruth. Methods with inaccurate anchors \cite{chai2021point,mao2021voxeltransformer} have poor performance in detecting large objects such as vehicles though they can have reasonable performance on pedestrians. One way to solve this problem is to have a two-stage model to refine the boxes \cite{mao2021voxeltransformer,shi2020pv} which greatly improves the detection accuracy. CenterNet-style detection methods \cite{ge2020afdet,yin2021centerpoint,sun2021rsn} strive to define anchors in the center of the groundtruth boxes only which enforces distributions of closer to zero mean and smaller variance. However, when detecting objects directly from sparse features (e.g. features from PointNet, Submanifold convolutions, sparse Transformers), there are not necessarily features  close to the object centers. To alleviate this issue,  \cite{sun2021rsn} applies normal sparse convolutions to insert points in the convolution output; \cite{singlestride21} scatters the sparse features to a dense BEV grid and runs dense convolutions to expand features to missing positions. These methods are expensive. In this paper, we propose a voxel diffusion module to address this issue efficiently in a scalable way by segmenting and diffusing foreground voxels to their nearby regions as described in \S\ref{sec:vd}.

Extensive experiments are conducted on the challenging Waymo Open Dataset \cite{sun2020scalability} to show state of the art results of {\ourmethod} on 3D object detection. 
We summarize our contributions as follows:

\begin{itemize}
\item We propose a hierarchical Sparse Window Transformer ({\ourmethod}) backbone for 3D representation learning. Its flexible receptive fields and multi-scale features make it suitable for different self-driving tasks like object detection and semantic segmentation.
\item We propose a generic voxel diffusion module to address the unique challenge of anchor placement in 3D object detection from sparse features.
\item We conduct extensive experiments on Waymo Open Dataset \cite{sun2020scalability} to demonstrate the state of the art performance of our {\ourmethod} model. 

\end{itemize}  

\section{Related Work}
 
\subsection{3D object detection}
As one of the most important tasks in autonomous driving, 3D object detection has been extensively studied in prior works. Early works like PointNet~\cite{qi2017pointnet} and PointNet++~\cite{qi2017pointnet++} directly apply multilayer perceptions on individual points, but it is difficult to scale them to large point clouds with good accuracy. The current mainstream 3D object detectors often convert point clouds into bird eye view 3D  ~\cite{zhou2018voxelnet} or 2D voxels~\cite{lang2019pointpillars} (2D voxels are also referred as pillars), where each voxel aggregates the information from points it contains. In this way, regular 2D or 3D convolutional neural networks can be applied to process these bird-eye-view representations. The pseudo image of voxels also makes it easier to reuse the rich research advancements in 2D object detection, such as two-stage or anchor-based detection heads~\cite{yin2021centerpoint}. The downside is that the pseudo image of voxels grows cubically/quadratically with the voxelization granularity and detection range, not to mention that many of the voxels are effectively empty. Therefore, another type of approach is to perform 3D object detection without voxelization. This includes methods that detect objects from the perspective view~\cite{meyer2019lasernet,chai2021point,fan2021rangedet}, or lookup nearest neighbors for each point~\cite{ngiam2019starnet}. However, the detection accuracy is typically inferior to the voxelization route.

To have the best of both worlds, recent approaches~\cite{yan2018second,sun2021rsn,shi2020pv} start to explore multi-view approaches and make use of \emph{sparse} convolutions on the \emph{voxelized} point cloud. For example, the recent range sparse net (RSN~\cite{sun2021rsn}) adopts a two-step approach, where the first step performs class-specific segmentation on the range image view, and the second step applies sparse 3D convolutions on the voxel view for specific classes. However, submanifold sparse convolutions cannot connect features that are sparsely disconnected without adding normal sparse convolutions and striding, and they often require heavily optimized customized ops to be efficient on modern accelerators. 

Our work aims to learn the 3D representations from sparse point clouds without using any dense or sparse convolutions. Instead, we resort to a hierarchical Transformer to achieve our goal.

\subsection{Transformers}

Transformers~\cite{vaswani2017attention} have shown great success in natural language processing~\cite{devlin2018bert}. Recently, researchers have brought this architecture to computer vision~\cite{bello2019attention,ramachandran2019stand,wang2018non,dai2021coatnet}.
ViT~\cite{dosovitskiy2020image} partitions images into patches, which greatly advanced the use of Transformers for image classification.
Swin Transformer~\cite{swin21} further demonstrated better ways to fuse contextual information through window shifting and hierarchy, and also generalized to other tasks such as segmentation and detection.

Interestingly, Transformers are naturally suitable for sparse point clouds, because they can take any length of sequences as inputs and do not require dense 2D/3D image representations. Therefore, recent works have attempted to adopt Transformers for 3D representation learning, but they are primary developed for object scans and indoor applications~\cite{zhao2021point,engel2021point,misra2021end,pan20213pointformer}.
Voxel Transformer~\cite{mao2021voxeltransformer} is the submanifold sparse convolution~\cite{graham2017submanifold} counterpart in the Transformer world, by replacing the convolution kernel with attention. Its irregular memory access pattern is computationally inefficient, and its accuracy is worse than state of the art methods. Recently, SST~\cite{singlestride21} proposes a single-stride transformer for 3D object detection and achieved impressive results on Waymo Open Datasets especially for pedestrian object detection. However, due to its single stride nature, SST has a limited receptive field and thus has difficulty dealing with large objects, making it ineffective in important tasks like large vehicle detection, large object segmentation (e.g. buildings), lane detection, and trajactory prediction. It needs to scatter features to a dense BEV grid to run several dense convolutions which limits its scalability. It is also computationally expensive as it needs to run many layers of transformers on the high resolution feature map which limits its applications in realtime systems.

Our work is inspired by window-based Transformers (e.g., SwinTransformer~\cite{swin21}) in the sense that we also adopt the hierarchical window-based Transformer backbone, but to address the unique challenges of 3D sparse point clouds, we propose several novel techniques such as the improved SWFormer blocks, multi-scale feature fusion, and voxel diffusion.


\section{Sparse Window Transformer}

\subsection{Overall Architecture}

\begin{figure*}[t]
    \centering
    \includegraphics[width=0.8\columnwidth]{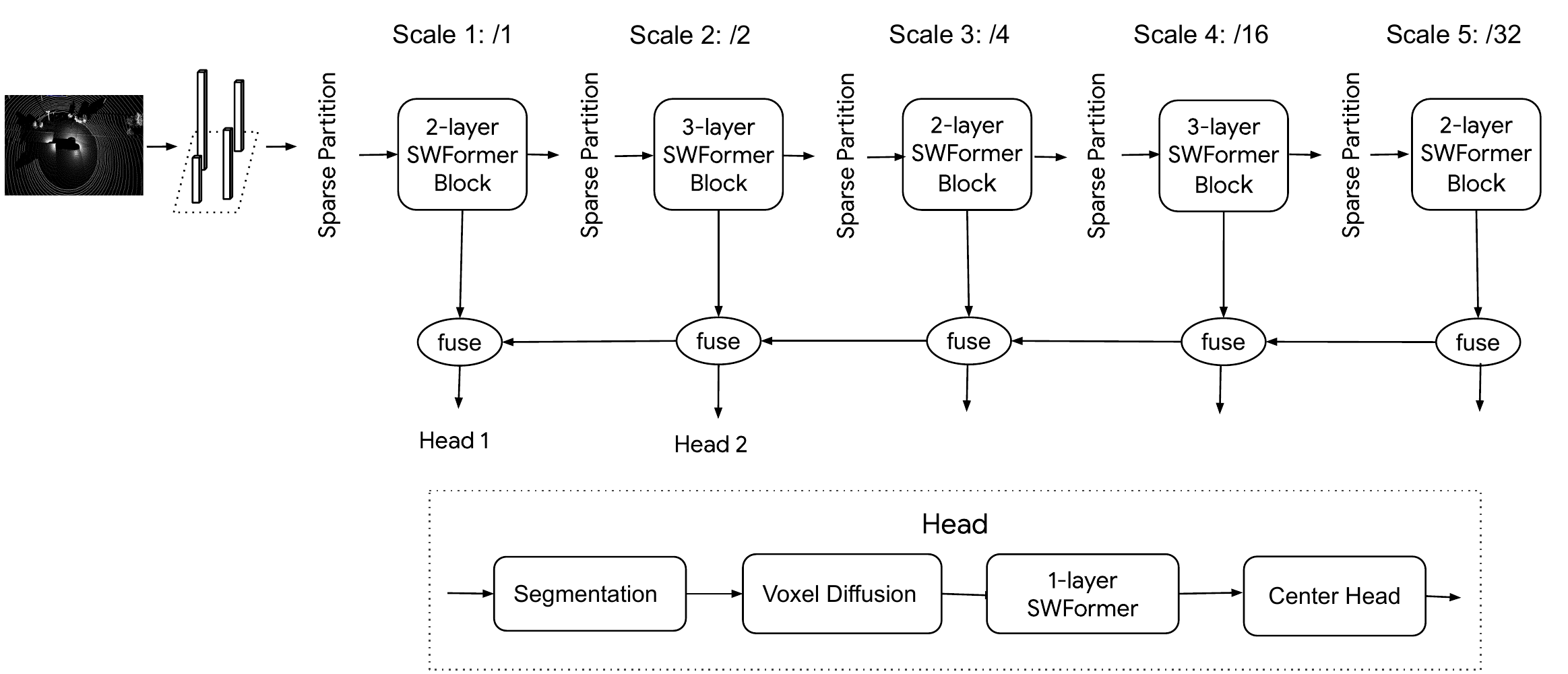}
    \caption{Overview of {\ourmethod} model architecture. Given a sparse point cloud, we first perform voxelization to generate a grid of 2D voxels. These voxels are then processed with a 5-scale sequence of hierarchical {\ourmethod} blocks (\autoref{fig:swformerblock}), with strides $\{1, 2, 4, 16, 32\}$. The output features are combined with a multi-scale feature fusion network (section \ref{sec:msff}). The fused features are fed to a head, which performs foreground segmentation and voxel diffusion (section \ref{sec:vd}),  and computes center net style classification and box regression loss (section \ref{sec:br}). Different object classes (e.g. vehicles and pedestrians) may use a separate head on different feature scales. 
    }
    \label{fig:arch}
\end{figure*}

{\ourmethod} is a pure Transformer-based model without any convolutions. \autoref{fig:arch} shows the overall network architecture: given a sequence of point cloud frames as inputs, each point is augmented with per-frame voxel features \cite{lang2019pointpillars} and an auxiliary frame timestamp offset \cite{sun2021rsn}. It uses dynamic voxelization \cite{zhou2019end} and a point net \cite{lang2019pointpillars,qi2017pointnet} based feature embedding net to get sparse voxel features. Note, our voxels are also referred as pillars in other works~\cite{lang2019pointpillars}.
These sparse voxels are then processed by a hierarchical sparse window Transformer network described in \S\ref{sec:hswt}. The resulting multi-scale features are then fused with a Transformer based feature fusion blocks. To address the unique challenge of detecting 3D boxes from sparse features, we first segment the foreground voxels and then apply a voxel diffusion module to expand foreground voxels to neighboring locations with pseudo voxels. In the end, we apply a center net \cite{yin2021centerpoint,sun2021rsn,zhou2019objects} style detection head to regress 3D boxes.

\subsection{Hierarchical Sparse Window Transformer Encoder}
\label{sec:hswt}

\begin{figure*}
    \centering
    \includegraphics[width=0.5\linewidth, trim={10, 210, 50, 0}, clip]{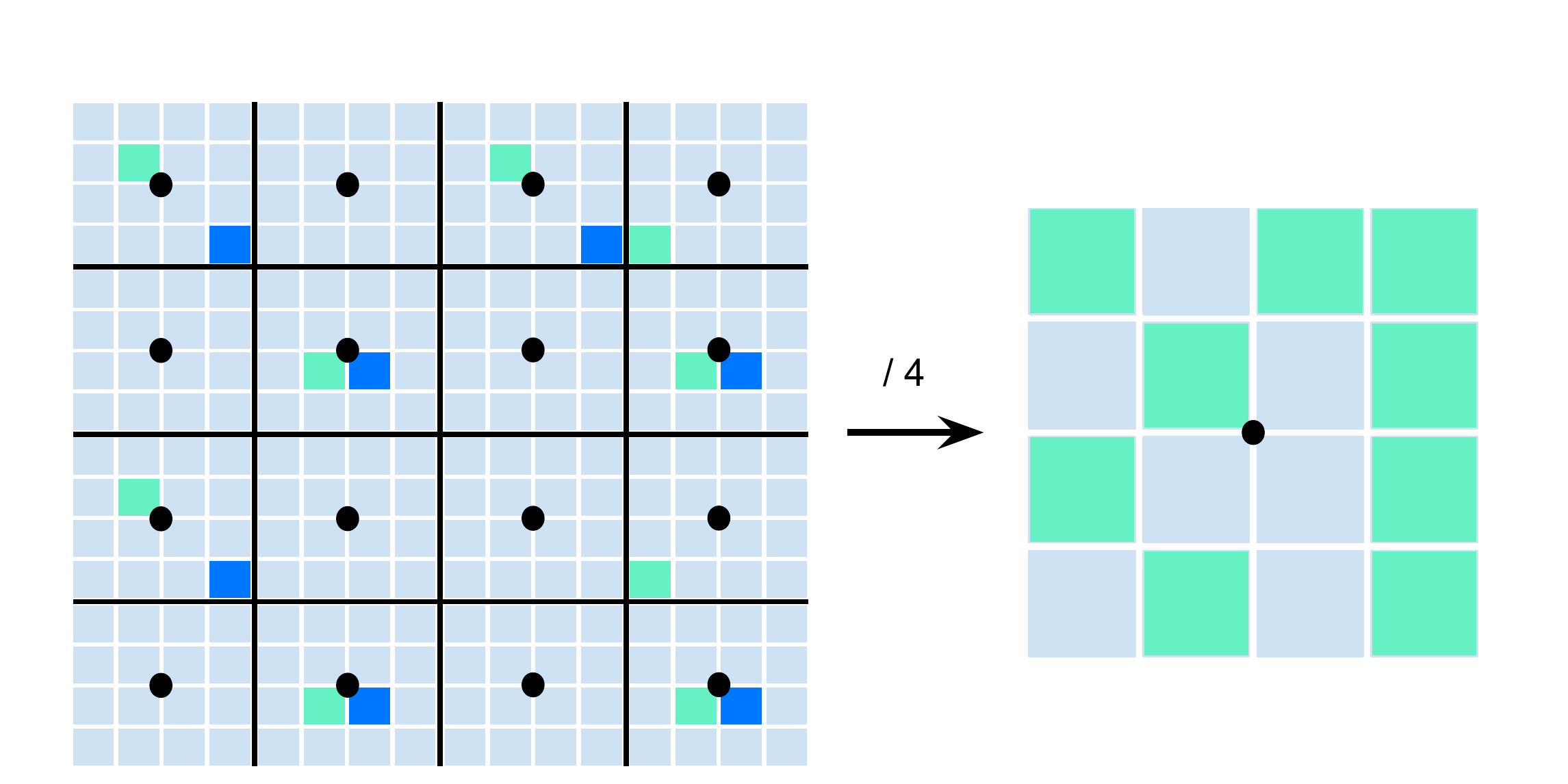}
    \caption{Strided Sparse Window Partition. Left shows a grid of 16x16 BEV voxels, where grey voxels are empty and others are non-empty. Right shows the results of stride-4 window partition, leading to a grid of 4x4  voxels. For each striding window, it picks the nearest neighbor non-empty voxel feature (light green) from the center (black dot) with any deterministic rule to break ties; if all voxels are empty in the striding window, then the corresponding voxel after striding is also empty. Best viewed in color.}
    \label{fig:win_partition}
\end{figure*}

A key concept of our {\ourmethod} is the \emph{sparse window} in the birds eye view. After points are converted to a grid of 2D voxels on bird eye view, the voxel grid is further partitioned into a list of non-overlapping windows with fixed size $H \times W$ (e.g., $10\times10$), similar to Swin Transformer\cite{swin21}; however, since points are often sparse, many voxels are empty with no valid points. Therefore, the number of non-empty voxels in each window may vary from 0 to $HW$. As we will explain later, all non-empty voxels within the same window will be flattened to a single variable-length sequence and fed into Transformer layers. In practice, these variable-length sequences prevent us from batch training, causing lower training efficiency. To solve this issue, we borrow a widely used ideas from natural language processing \cite{vaswani2017attention,bert19} and recent works \cite{singlestride21}, which group these sparse windows into different buckets based on their sequence lengths. Concretely, we divide sparse windows into at most $k$ buckets $\{B_0, B_1, ..., B_k\}$,  where windows in $B_i$ are always padded to a maximum sequence length of $HW/2^i$. All padded tokens are masked in Transformer layers.

Based on the aforementioned sparse windows, our encoder adopts hierarchical Transformers to process the inputs and produce a list of multi-scale BEV features. As shown in \autoref{fig:arch}, each scale starts with a sparse window partition layer followed by a multi-layer SWFormer block.

\textbf{Sparse Window Partition:} We divide the BEV voxels into non-overlapping windows with fixed size $H\times W$, which are then grouped into buckets $\{B_0, B_1, ..., B_k\}$. For each bucket $B_i$, we flatten all voxels within the same window into a sequence and zero-pad the sequence length to $HW/2^i$. These sequences are then batched and fed to the Transformer blocks, where the self-attention shares the keys and values for all query voxels coming from the same window \cite{swin21}.
Since {\ourmethod} processes inputs in a hierarchical fashion with multiple feature scales, we need to apply strided window partitions at the beginning of each scale. The strided window partition is similar to traditional strided convolutions, except that it always picks the closest voxel to the center of the window with deterministic rules to break ties. Notably, no max or average pooling operations are applied because they are not friendly to sparse implementations. \autoref{fig:win_partition} illustrates an example of a stride-4 window partition.

\begin{figure}[!h]
    \centering
    \includegraphics[width=0.6\columnwidth, trim={50, 100, 50, 50}, clip]{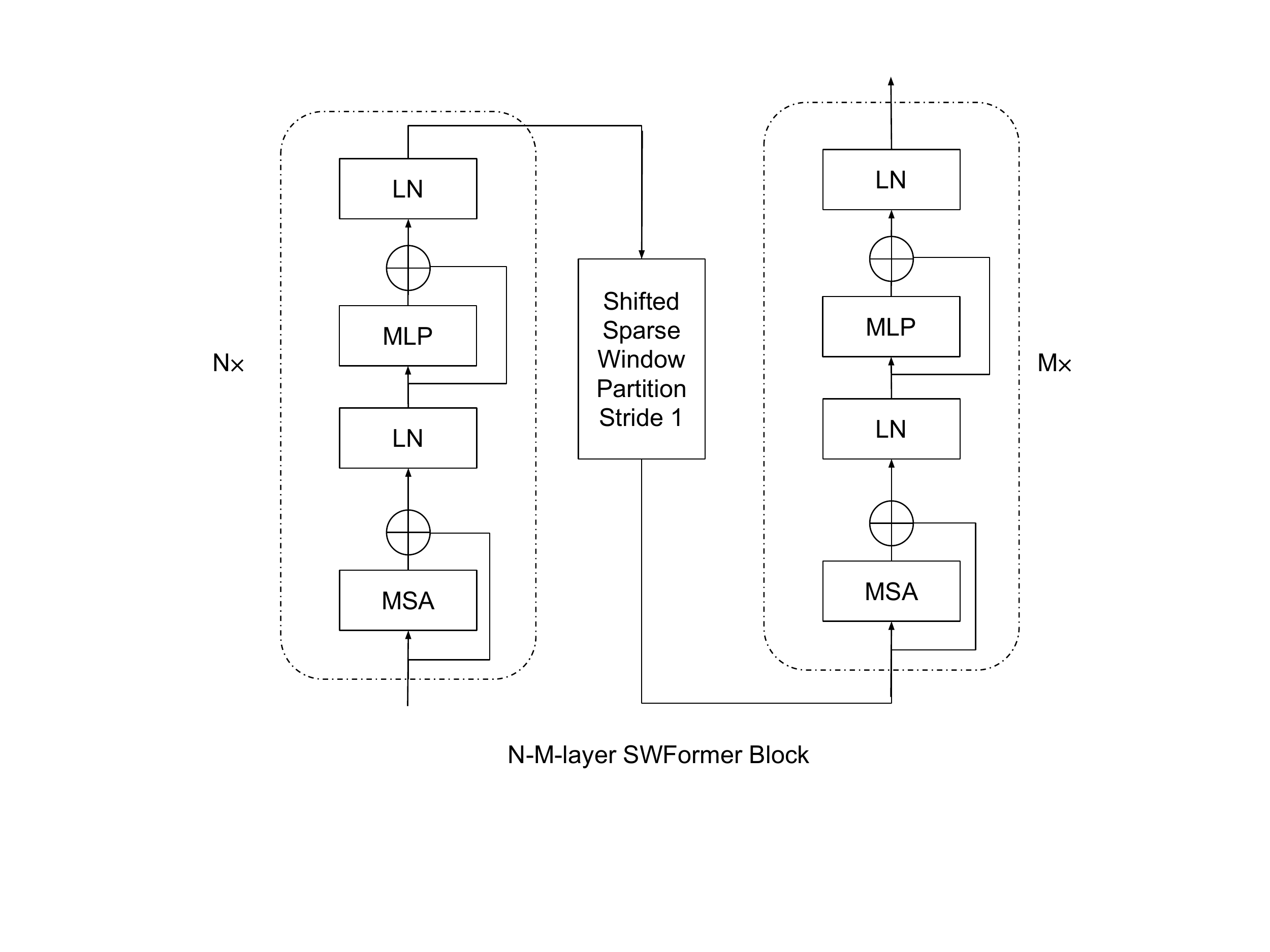}
    \caption{Sparse Window Transformer Block. Given a sequence of sparse features, it first applies a multi-head self-attention (MSA) on all valid voxel within the same window, followed by a MLP and layer norm. After repeating the Transformer layer $N$ times, it performs a shifted sparse window partition to re-generate the sparse windows, and then process the shifted windows with another $M$ Transformer layers. If $N$ and $M$ are the same, we name it as N-layer SWFormer block for simplicity.}
    \label{fig:swformerblock}
\end{figure}

\textbf{Sparse Window Transformer block:} Transformer\cite{vaswani2017attention}  is inherently suitable for sparse point clouds, as it does not require the dense 2D/3D inputs as in convolutional networks; unfortunately, due to the quadratic complexity of self-attention with respect to the input sequence length, it is prohibitively expensive to feed the whole point cloud (with millions of points) or voxel features (with tens of thousands valid voxels) as a single input sequence to Transformer. In this paper, we adopt the idea of Swin Transformer~\cite{swin21}: the sparse BEV voxels are first partitioned into windows, and Transformer is applied to each window separately. To increase the receptive field and connect the features across windows, SwinTransformer uses a window shifting technique to re-partition the window for every layer of Transformer. However, as we are operating on sparse voxel features, such shift-window operation is memory-read/write intensive, especially for matrix-optimized accelerators like TPUs. To alleviate this problem, we propose to limit the shift-window operation to once per stride rather than per layer. \autoref{fig:swformerblock} shows the detailed architecture of a {\ourmethod} block: it largely follows the same style of SwinTransformer to perform self-attention within a local window, except it only performs shift-window operation once in the middle.
Formally, our {\ourmethod} block can be described as follows:
\begin{align}
    & {\bf z}^0 = [{\bf x}; \text{ mask}_z] + \text{PE}_{z} & \\
    &{{\hat{\bf{z}}}^{l}} = \text{LN}\left( {\bf{z}}^{l - 1} + {\text{MSA}( {{{\bf{z}}^{l - 1}}} )}   \right)  & l = 1...N \nonumber\\
    &{{\bf{z}}^l} = \text{LN}\left( {{\hat{\bf{z}}}^{l}} + {\text{MLP} ({{{\hat{\bf{z}}}^{l}}} )} \right)  & l = 1 ... N \nonumber\\
    & {\bf u}^0 = [{\text{shift-window}(\bf z}^N); \text{ mask}_u] + \text{PE}_{u} & \\
    &{{\hat{\bf{u}}}^{l}} = \text{LN}\left( {\bf u }^{l - 1} + {\text{MSA}( {{{\bf{u}}^{l - 1}}} )}   \right)  & l = 1...M \nonumber\\
    &{{\bf{u}}^l} = \text{LN}\left( {{\hat{\bf u}}^{l}} + {\text{MLP} ({{{\hat{\bf{u}}}^{l}}} )} \right)  & l = 1 ... M \nonumber
    \label{eq.SWFormer}
\end{align}

\noindent where ${\bf x}$ is the input features after sparse window partition, $\text{mask}_z$ is the mask for input padding,  $\text{PE}_z$ is the positional encoding. The process contains two stages: (1) the first stage applies $N$ Transformer layers to ${\bf z}^0$ and output ${\bf z}^N$. Each Transformer layer consists of a standard multi-head self-attention (MSA) and multilayer perceptron (MLP), but slightly different from the standard version, here we adopt the post-norm scheme where layer norm (LN) is added after MSA and MLP. For simplicity, we use the standard sine/cosine absolute positional encoding in this paper. (2) The second stage first applies window-shift to ${\bf z}^N$, and adds the updated $\text{mask}_u$ and positional encoding $\text{PE}_u$ based on ${\bf z}^N$; afterwards, $M$ Transformer layers are added to process ${\bf u}^0$ and generate the final output ${\bf u}^M$. Notably, each SWFormer block has $N+M$ Transformer layers but only one window-shift operation.

By restricting window-shift operations, our SWFormer block is more efficient than the conventional Swin Transformer; however, it also limits the receptive field, since each Transformer layer is only applied to a small window. To address this challenge, {\ourmethod} is designed as a hierarchical network with multiple scales, where the strides are gradually increased: for simplicity, this paper uses strides $\{1, 2, 4, 16, 32\}$ for the five scales. For each scale, we always keep the window size fixed (e.g., $10\times 10$); however, as the later scales have larger strides, the same window in later scales will cover much larger area. As an example, for the last scale with stride 32, a $10\times 10$ window would cover $320\times 320$ area on the original BEV voxel grid, and a single window-shift would connect all features within an area as large as $480\times 480$.

\subsection{Multi Scale Feature Fusion}
\label{sec:msff}

\begin{wrapfigure}{r}{0.35\columnwidth}
    \centering
    \includegraphics[width=0.4\columnwidth]{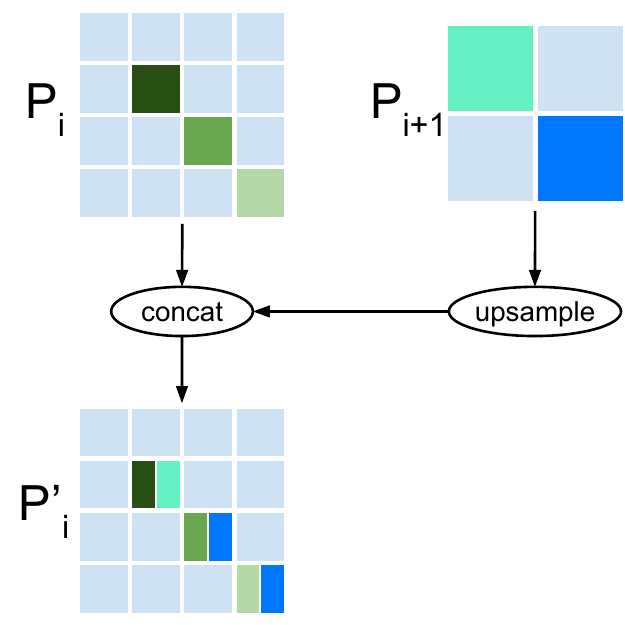}
    \caption{\footnotesize Feature Fusion. Feature $P_{i+1}$ is upsampled and concatenated with $P_i$ to generate $P'_i$ and the final $P_i$. During upsampling, we only duplicate $P_{i+1}$ features to locations that are non-empty in $P_i$.}
    \label{fig:featurefusion}
\end{wrapfigure}

Inspired by feature pyramid network (FPN \cite{lin2017feature}), {\ourmethod} adopts Transformer-based multi-scale feature network to effectively combine all features from the hierarchical Transformer encoder.  \autoref{fig:featurefusion} shows the overall architecture of the feature network: given a list of encoder features $\{P_0, P_1, .. P_5\}$, it iteratively fuses $(P_{i+1}, P_i)$ from large-stride $P_5$ to small-stride $P_0$.
Formally, our feature fusion process can be described as:

\begin{align}
    & \hat{P}_5 = P_5 & \\
    & \hat{P}_i = \text{SWFormer}(\text{Concat}(P_i, \text{Upsample}(\hat{P}_{i+1})))  & i = 0, ..., 4
\end{align}

\noindent Starting from the last feature map $P_5$, we first upsample it to have the same stride as $P_4$ such that they can be concatenated into a single feature map; afterwards, we simple apply a 1-layer {\ourmethod} block to process the concatenated feature and generate the new $\hat{P}_4$. The process is iterated until all fused features $\{\hat{P}_0, ..., \hat{P}_5\}$ have been generated, which have the same strides as $\{P_0, ..., P_5\}$ features. The fused features are further used in voxel diffusion and box regression as described in the following sections.

One challenge in sparse upsamping is that one cannot naively duplicate the feature to all upsampled locations (like commonly done in dense upsampling), which will cause unnecessary excessive feature duplication and significantly reduce the sparsity.  In this paper, we  restrict features in $P_{i+1}$ to only duplicate to locations that have non-empty features in $P_i$, as shown in \autoref{fig:featurefusion}. In this way, we can ensure $\hat{P}_i$ has the same sparsity as $P_i$.

\subsection{Voxel Diffusion}
\label{sec:vd}

\begin{figure}
    \centering
    \includegraphics[width=0.9\columnwidth, trim={20, 220, 80, 20}, clip]{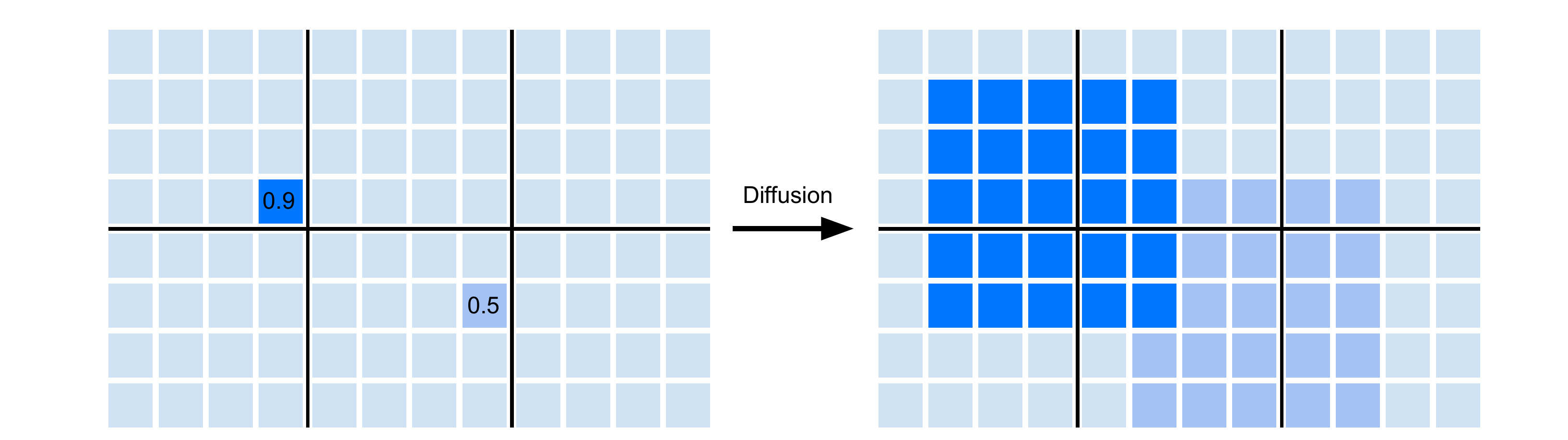}
    \caption{Voxel Diffusion. After foreground segmentation, each voxel receives a segmentation score $s \in [0, 1]$. All voxels with scores greater than a threshold $\gamma=0.05$ are scattered to a dense BEV grid, and then we apply a $k\times k$ max pooling on the dense BEV grid to expand valid voxel features to their neighboring locations where k is set to 5 in this example. (Left) before diffusion, there are only two foreground voxels with segmentation scores \{0.5, 0.9\} greater than $\gamma$; (Right) after voxel diffusion, 47 voxels become valid. Best viewed in color.}
    \label{fig:voxel_diffusion}
\end{figure}

To detect 3D objects from sparse voxel features, a unique challenge is that there might be no valid voxel feature near object centers which are the best positions to place implicit \cite{yin2021centerpoint} or explicit anchors \cite{ren2016faster}. Prior works have attempted to resolve this issue by: 1) second-stage box refinement \cite{shi2020pv}, 2) sparse convolutions \cite{sun2021rsn} or coordinate refinement \cite{pan20213pointformer} that can expand features to empty voxels close to the object centers, 3) scattering sparse voxel features to dense and applying dense convolutions \cite{singlestride21}.  In this paper, we propose a novel \emph{voxel diffusion} module to effectively and efficiently address this challenge.

Voxel diffusion is based on two simple ideas: First, we segment all foreground voxels by jointly performing foreground/backgrond segmentation, thus effectively filtering out the majority of background voxels. Second, we expand all foreground voxels by zero-initializing their features into neighboring locations with a simple $k\times k$ max pooling operations on the dense BEV grid, where  $k$ is the detection head specific diffusion factor to control the magnitude of expansion. The diffused voxel features are further connected and processed  with a few Transformer layers. Combining these two ideas, we can simultaneously keep voxel features sparse (by filtering out background voxels) and features filled (by voxel diffusion) for voxels closer to the object center. \autoref{fig:voxel_diffusion} illustrates an example of voxel diffusion.

Our foreground segmentation is jointly trained with object detection. Specifically, for each voxel, we assign a binary groundtruth label: 0 (background, voxel does not overlap with any objects) and 1 (foreground, voxel overlaps with at least one object). The foreground segmentation is trained with a two-class focal loss \cite{lin2017focal} for each object class $c$:

\begin{equation}
    L_\text{seg}^{c} = \frac{1}{N} \sum_i{L_i}
\end{equation}

\noindent where $N$ is the total number of valid voxels and $L_i$ is the focal loss for voxel $i$. At inference time, we keep voxels as foreground if their foreground scores are greater than a threshold $\gamma$.

\subsection{Box Regression}
\label{sec:br}
{\ourmethod} follows \cite{sun2021rsn} to use a modified CenterNet~\cite{zhou2019objects,ge2020afdet,sun2021rsn,yin2021centerpoint} head to regress boxes from voxel features. The heatmap loss is computed as a penalty-reduced focal loss \cite{zhou2019objects,lin2017focal} per object class.
\begin{align}
\label{eq:hm_loss}
\begin{split}
    L_{\textrm{hm}}^{c} = -\frac{1}{N}\sum_{i}\{
    (1 - \Tilde{h}_{i})^\alpha\log(\Tilde{h}_{i})I_{h_{i} > 1 - \epsilon} + \\ (1-h_{i})^\beta\Tilde{h}_{i}^\alpha\log(1-\Tilde{h}_{i})I_{h_{i} \leq 1 - \epsilon}\},
\end{split}
\end{align}
where $\Tilde{h}_{i}$ and $h_{i}$ are the predicted and ground truth heatmap values for object class $c$ respectively at voxel $i$. $N$ is the number of boxes in class $c$. We use $\epsilon = 1e-3$, $\alpha=2$ and $\beta=4$ in all experiments, following \cite{zhou2019objects,law2018cornernet,sun2021rsn}. {\ourmethod} parameterize 3D boxes as $\boldsymbol{b} = \{d_x, d_y, d_z, l, w, h, \theta\}$ where $d_x, d_y, d_z$ are the box center offsets relative to the voxel centers. $l, w, h, \theta$ are box length, width, height and box heading. We follow \cite{sun2021rsn} to apply a bin loss \cite{shi2019pointrcnn} to regress heading $\theta$, smooth L1 to regress other box parameters, and an IoU loss \cite{zhou2019iou} to improve overall box accuracy on the voxels with ground truth heatmap values above a threshold $\delta_1$.
\begin{align}
\label{eq:bin_loss}
L_{\theta_i}^c &= L_{bin}(\theta_i, \Tilde{\theta}_i), \\
L_{\boldsymbol{b_{i}} \backslash \theta_{i}}^c &= \textrm{SmoothL1}(\boldsymbol{b_{i}} \backslash \theta_{i} -  \boldsymbol{\Tilde{b_{i}}} \backslash \Tilde{\theta}_i), \\
L_{\textrm{box}} ^ c &= \frac{1}{N} \sum_i {(L_{\theta_i} + L_{\boldsymbol{b_i} \backslash \theta_{i}} + L_{\textrm{iou}_{i}}) I_{h_i > \delta_1}},
\end{align}
where $\Tilde{b}_i$, $b_i$ are the predicted and ground truth box parameters respectively, $\Tilde{\theta}_i$, $\theta_i$ are the predicted and ground truth box heading respectively.

The net is trained end to end with the total loss defined as
\begin{equation}
\label{eq:total_loss}
L  = \sum_{c}(\lambda_1 L_{\textrm{seg}}^c + \lambda_2 L_{\textrm{hm}}^c + L_{\textrm{box}}^c)
\end{equation}

When decoding prediction boxes, we first filter voxels with heatmap less than a threshold $\delta_{2}$, then run max pool on the heatmap to select boxes corresponding to the local heatmap maximas without any non-maximum-suppression.

\section{Experiments}
\label{expt}
We describe the {\ourmethod} implementation details, and demonstrate its efficiency and accuracy in multiple experiments. Ablation studies are conducted to understand the importance of various design choices.

\subsection{Waymo Open Dataset}
Our experiments are primary based on the challenging Waymo Open Dataset (WOD)~\cite{sun2020scalability}, which has been adopted in many recent state of the art 3D detection methods \cite{shi2020pv,yin2021centerpoint,sun2021rsn,singlestride21,qi2021offboard}. The dataset contains 1150 scenes, split into 798 training, 202 validation, and 150 test. Each scene has about 200 frames, where each frame captures the full 360 degrees around the ego-vehicle. The dataset has one long range LiDAR with range capped at 75 meters, four near range LiDARs and five cameras. {\ourmethod} uses all five LiDARs in the experiments.

\subsection{Implementation Details}
We normalize intensity and elongation in the raw point cloud with the $\textrm{tanh}$ function.
The dynamic voxelization uses $0.32m$ voxel size in $x$, $y$ and infinite size in $z$. During training, we ignore all ground truth boxes with fewer than five points inside. The voxel feature embedding net has two layers of MLPs with channel size of 128. All of the transformer layers have channel size of 128, 8 heads, and inner MLP ratio of 2. We also use stochastic depth~\cite{droppath16} with survival probability 0.6. The segmentation cutoff $\gamma$ in \S\ref{sec:vd} is set to $0.05$. The heatmap threshold $\delta_{1}$, $\delta_{2}$ are set to $0.2$, $0.1$ respectively for both vehicle and pedestrian heads. For training efficiency, we cap the number of regression targets in each frame by 1024 for vehicle and 800 for pedestrian sorted by ground truth heatmap values. $\lambda_1$, $\lambda_2$ are set to 200 and 10 in Eq.~\ref{eq:total_loss}.

\textbf{Data augmentation.} We have adopted the several popular 3D data augmentation techniques described in \cite{cheng2020ppba} during training: randomly rotating the world by yaws uniformly chosen from $[-\pi, \pi]$ with probability $0.74$, randomly flipping the world along y-axis with probability $0.5$, randomly scaling the world with scaling factor uniformly chosen within $[0.95, 1.05)$, randomly dropping points with probability of $0.05$.

\textbf{Training and Inference.} The {\ourmethod} models are trained end-to-end with 32 TPUv3 cores using the Adam optimizer \cite{kingma2014adam} for a total number of 128 epochs with an initial learning rate set to 1e-3. We apply cosine learning rate decay and 8 epoch warmup with initial warmup learning rate set to 5e-4.

\subsection{Main Results}

We measured the detection results using the official WOD detetion metrics: BEV and 3D  average precision (AP), heading error weighted BEV, and 3D average precision (APH) for L1 (easy) and L2 (hard) difficulty levels \cite{sun2020scalability}. The official metrics used to rank in the leaderboard uses IoU cutoff of 0.7 for vehicle, 0.5 for pedestrian. We report additional AP results at IoU of 0.8 for vehicle, 0.6 for pedestrian. Large vehicles that have max dimension greater than 7 meters are also reported. 
\autoref{vehicle_ped_result} reports the main results on validation set,  \autoref{additional_validation_results} reports additional results for high IoU and large vechiels on the validation set, and \autoref{test_result} shows the test set results by submitting our predictions to the official test server. Results from methods with test time augmentation or emsemble are not included.

As shown in \autoref{vehicle_ped_result}, {\ourmethod} achieves new state-of-the-art results for vehicle detection on the WOD \textit{validation set}: it has 1.5 APH/L2 higher than the prior best single-stage model RSN~\cite{sun2021rsn}. {\ourmethod} even outperforms the prior best performing two-stage method PVRCNN++\cite{shi2021pv++} by 0.42 APH/L2. Importantly, {\ourmethod} performs very well at detecting large vehicles, 6.35 AP/L2 higher than the prior art of RSN \cite{sun2021rsn} as shown in \autoref{additional_validation_results}. {\ourmethod} slightly outperforms the state of the art single stage method SST\_3f \cite{singlestride21} by 0.12 APH/L2. Notably, the single frame single stage {\ourmethod}\_1f also outperforms all prior single frame methods.

We have compiled the model with XLA~\cite{xla} and ran inference for the 15th frame in scene 8907419590259234067\_1960\_000\_1980\_000 that has 68 vehicles and 69 pedestrians on a Nvidia T4 GPU. The latency is 43ms, more efficient than the popular realtime detector PointPillars \cite{lang2019pointpillars} which takes about 100ms on the same GPU with our own implementation. With fused transformer GPU kernels and optimized GPU sparse window partition operations, the latency can be further reduced to 20ms.

\begin{table}[h]
\caption{WOD \textit{validation set} results. \dag~is from~\cite{sun2021rsn}. Top methods are highlighted. Top one-frame (cyan), single-stage (blue) are colored. TS: two-stage. BEV: BEV L1 AP.}
\begin{center}
 \resizebox{0.85\textwidth}{!}{
\begin{tabular}{l|c|ccc|ccc}
\toprule
\multirow{2}{*}{Method} &
\multirow{2}{*}{TS} &
\multicolumn{3}{c|}{AP/APH Vehicle} &
\multicolumn{3}{c}{AP/APH Pedestrian} \\
& & 3D L1 & 3D L2 & BEV & 3D L1 & 3D L2 & BEV \\
\midrule
PVRCNN++ \cite{shi2021pv++} & \cmark & 79.3/78.8& {70.6/70.2} & -& 81.8/76.3 & 73.2/68.0  & -\\
VoTr-TSD\cite{mao2021voxeltransformer} & \cmark & 75.0/74.3 & 65.9/65.3 & - & - & - &- \\
SST\_TS\_3f \cite{singlestride21} & \cmark &78.7/78.2&70.0/69.6&-&\textbf{83.8/80.1}&\textbf{75.9/72.4} & -  \\
CenterPoint\_TS \cite{yin2021centerpoint} & \cmark & 76.6/76.1 &68.9/68.4 & - & 79.0/73.4 & 71.0/65.8 &  - \\
PointPillars \cite{lang2019pointpillars} \dag  & \xmark & 63.3/62.7 & 55.2/54.7 & 82.5 & 68.9/56.6  &  60.0/49.1 & 76.0 \\
MVF++\_1f \cite{qi2021offboard} & \xmark & 74.6/- & - & 87.6 & 78.0/-  & - & 83.3 \\
RSN\_1f \cite{sun2021rsn} & \xmark & 75.1/74.6 & 66.0/65.5 & 88.5 &77.8/72.7 &68.3/63.7 &83.4 \\
RSN\_3f \cite{sun2021rsn} & \xmark & {{78.4/78.1}} & {{69.5/69.1}} & {{91.3}} & 79.4/76.2 & 69.9/67.0 & 85.0  \\
SST\_1f \cite{singlestride21} & \xmark &74.2/73.8&65.5/65.1&-&78.7/69.6&70.0/61.7&- \\
SST\_3f \cite{singlestride21} & \xmark &77.0/76.6 &68.5/68.1&-&{82.4/78.0} & {\blue{75.1}}/70.9 \\
\midrule
SWFormer\_1f (Ours) & \xmark & \cyan{77.8/77.3} & \cyan{69.2/68.8} & \cyan{91.7} & \cyan{80.9/72.7} & {\cyan{72.5}}/\cyan{64.9} & \cyan{86.1} \\
SWFormer\_3f (Ours) & \xmark & \textbf{79.4/78.9} & \textbf{71.1/70.6} & \textbf{92.6} & \blue{82.9/79.0} & 74.8/\blue{71.1} & \blue{87.5} \\
\bottomrule
\end{tabular}
}
\end{center}
\label{vehicle_ped_result}
\end{table}

Table~\ref{test_result} shows vehicle and pedestrian detection result comparison with published results on the WOD \textit{test set}, which shows \ourmethod outperforms all previous single-stage or two-stage methods on the official ranking method mAPH/L2. 

\begin{table}
\caption{Additional WOD \textit{validation set} results. Top methods are highlighted.}
\begin{center}
\resizebox{0.65\textwidth}{!}{
\begin{tabular}{l|c|c|c|c}
\toprule
\multirow{2}{*}{Method} &
\multicolumn{3}{c|}{Vehicle L1 AP} &
\multicolumn{1}{c}{Pedestrian L1 AP} \\
& 3D IoU=0.8 & BEV Large & 3D Large & 3D IoU=0.6  \\
\toprule
MVF++\_1f \cite{qi2021offboard} & 43.3 & - & - & 56.0 \\
RSN\_3f \cite{sun2021rsn}& 46.4 & 53.1 & 45.2 & - \\
\midrule
SWFormer\_3f (Ours) &\textbf{47.5} & \textbf{60.1} & \textbf{51.5} & \textbf{62.1} \\
\bottomrule
\end{tabular}
}
\end{center}
\label{additional_validation_results}
\end{table}


\begin{table}[!h]
\centering
\caption{WOD \textit{test set} results. \dag~is from~\cite{sun2021rsn}. Top methods are highlighted. mAPH/L2 is the official ranking metric on the WOD leaderboard. TS is short for two-stage.}
\centering
 \resizebox{0.85\textwidth}{!}{
\begin{tabular}{l|c|c|cc|ccc}
\toprule
\multirow{2}{*}{Method} &
\multirow{2}{*}{TS} &
\multirow{1}{*}{mAPH} &
\multicolumn{2}{c}{Vehicle AP/APH 3D} &
\multicolumn{2}{c}{Pedestrian AP/APH 3D} \\
& & L2 & L1 & L2 & L1 & L2 \\
\midrule
CenterPoint~\cite{yin2021centerpoint} &\cmark & 69.1 &80.20/79.70 & 72.20/71.80 & 78.30/72.10 & 72.20/66.40 \\
SST\_TS\_3f~\cite{singlestride21} & \cmark &72.94&80.99/80.62 & 73.08/72.74 & 83.05/79.38 & 76.65/73.14 \\
PVRCNN++~\cite{shi2021pv++} & \cmark & 71.24&81.62/81.20  & 73.86/73.47 & 80.41/74.99 & 74.12/69.00 \\
P.Pillars~\cite{lang2019pointpillars} \dag & \xmark &55.10& 68.60/68.10 &  60.50/60.10 &  68.00/55.50 & 61.40/50.10 \\
RSN\_3f~\cite{sun2021rsn} & \xmark &69.70& 80.70/80.30 & 71.90/71.60 & 78.90/75.60 & 70.70/67.80 \\
\midrule
SWFormer\_3f (Ours) & \xmark &\textbf{73.36}& 82.89/82.49 & 75.02/74.65 & 82.13/78.13 & 75.87/72.07 \\
\bottomrule
\end{tabular}
}
\label{test_result}
\end{table}


\subsection{Ablation Study}

\noindent\textbf{Voxel diffusion} is one of the primary contributions of this paper. We study its impacts by varying the diffusion window size $k$ introduced in \S\ref{sec:vd}. The result in \autoref{abalation_diffusion} shows  the significance of voxel diffusion. Disabling voxel diffusion (i.e. setting $k=1$) results in 6.37 and 3.22 3D AP drop compared with $k=9$ on vehicle and pedestrian detection respectively. Increasing $k$ can slightly improve the detection accuracy especially on vehicle.

\noindent\textbf{Multi-scale feature} improves the model accuracy as shown in \autoref{abalation_study} especially going from one scale to two scales. The impact is larger on vehicle detection (+2.72 3D AP) than pedestrian detection (+1.15 3D AP). The 3-scale model has pretty close accuracy as the full 5-scale model. In practice, we can trade-off between accuracy and latency by adjusting the number of scales. Note that some autonomous driving tasks such as lane detection, behavior prediction require larger receptive field. The success of training a deep five-scale {\ourmethod} model shows its potential in those tasks.

\begin{table}
\caption{Impact of Voxel Diffusion. Compared to the baseline (window size = 1), our voxel diffusion improves accuracy, especially with large diffusion window size.}
\centering
 \resizebox{0.5\textwidth}{!}{
\begin{tabular}{l|cccc}
\toprule
\multirow{1}{*}{Diffusion Window Size}
& 1 & 3 & 5 & 9  \\
\midrule
Vehicle 3D AP/L1 &  72.13 & 78.07 &  78.58 &  78.50 \\
Pedestrian 3D AP/L1  & 79.23 & 82.28 & 82.44 &  82.45  \\
Vehicle BEV AP/L1  & 82.42 & 91.19 & 92.09  & 92.03 \\
Pedestrian BEV AP/L1  & 83.65  & 87.01  & 87.15 &  87.47 \\
\bottomrule
\end{tabular}
}
\label{abalation_diffusion}
\end{table}

\begin{table}
\caption{Impact of Multi-Scale and Window Shifting. Compared to single scale, multi-scale have much better accuracy. Window shifting is also important for performance. }
\centering
 \resizebox{0.6\textwidth}{!}{
\begin{tabular}{l||cccc||cc}
\toprule
\multirow{2}{*}{}
& \multicolumn{4}{c||}{Number of Scales}
& \multicolumn{2}{c}{Window Shift} \\
& 1 & 2 & 3 & 5 & \xmark & \checkmark \\
\midrule
Vehicle 3D AP/L1 & 74.96 & 77.68 & 78.88 & 79.36 & 76.74 & 79.36 \\
Pedestrian 3D AP/L1 & 81.24 & 82.39 & 82.19 & 82.91 & 81.19 & 82.91 \\
Vehicle BEV AP/L1 & 89.55 & 91.83 & 92.23 & 92.60 & 90.74 & 92.60\\
Pedestrian BEV AP/L1 & 86.48 & 87.30 & 87.13 & 87.54 & 86.46 & 87.54 \\
\bottomrule
\end{tabular}
}
\label{abalation_study}
\end{table}

\noindent\textbf{Window shifting} is introduced in SwinTransformer~\cite{swin21} to connect the features among windows. We have limited its usage to one per scale. What happens if we completely remove it? \autoref{abalation_study} shows clear accuracy drop especially on vehicles if the window-shift operations are removed from the {\ourmethod} blocks. This meets our intuition that it is important to keep one window shift operation per scale to make sure every voxel gets the similar receptive field in all directions.


\section{Conclusion}
This paper presents \emph{\ourmethod}, a scalable and accurate sparse window transformer-only model, to effectively learn 3D point cloud representations for object detection. Built upon window-based Transformers, it addresses the unique challenges brought by the sparse 3D point clouds, and proposes a bucketing-based  multi-scale Transformer neural network. {\ourmethod} takes full advantage of the sparsity of point clouds, and can effectively processes sparse windows of point clouds using pure Transformer layers without any convolutions. It also proposes a novel voxel diffusion module to further detect 3D objects from sparse features. Experiments show state-of-the-art results on the challenging Waymo Open Dataset.

\clearpage
%
%
\bibliographystyle{sty/splncs04}
\bibliography{cv}

\clearpage
\newpage

\appendix
 
\section{Window Shift}
\begin{figure*}
\centering
   \includegraphics[width=0.75\linewidth]{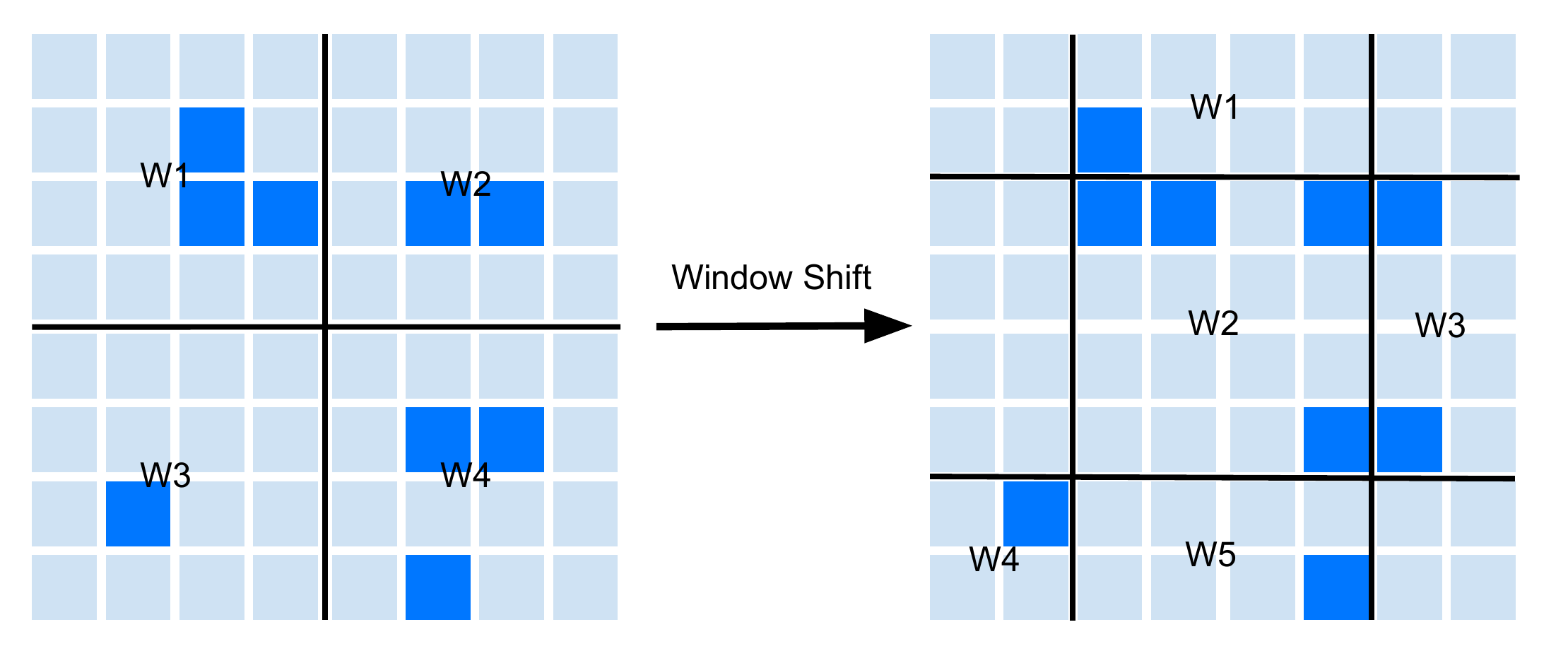}
   \caption{Sparse window shift. The dark blue cells are voxels with points. The light blue cells are voxels without points. Left shows a grid of $8\times8$ BEV voxels partitioned into 4 non-empty sparse windows with window size of $4\times4$. After window shift, it results in 5 non-empty sparse windows as shown on the right.}
\label{fig:window_shift}
\end{figure*}

The window shift operation (\autoref{fig:window_shift}) is implemented by adding offsets of half of the window sizes to the voxelized coordinates and then running the same window partition algorithm. We have proposed to limit the window shift operation per scale for efficiency of processing sparse inputs. We wondered what happens if we add more window shifts to the model. Will it impact model accuracy? We added one more window shift to the scale 1 and scale 2 respectively. It has slowed down our training by $10\%$. Surprisingly, it slightly decreased the model accuracy as shown in \autoref{more_shift}. Our hypothesis is that more shifts make the model harder to train when we do not need to rely on window shifts to increase receptive field. 

\begin{table}
\caption{Impact of adding more window shifts.}
\centering
\begin{tabular}{c|c|c}
\toprule
\multirow{1}{*}{More Window Shifts}
& Vehicle 3D AP/L1 & Pedestrian 3D AP/L1 \\
\midrule
\xmark &  79.36 & 82.91  \\
\cmark  & 79.17 & 82.36   \\
\bottomrule
\end{tabular}
\label{more_shift}
\end{table}

\section{Qualitative Results}
\begin{figure}[h!] 
\centering
   \includegraphics[width=0.8\linewidth]{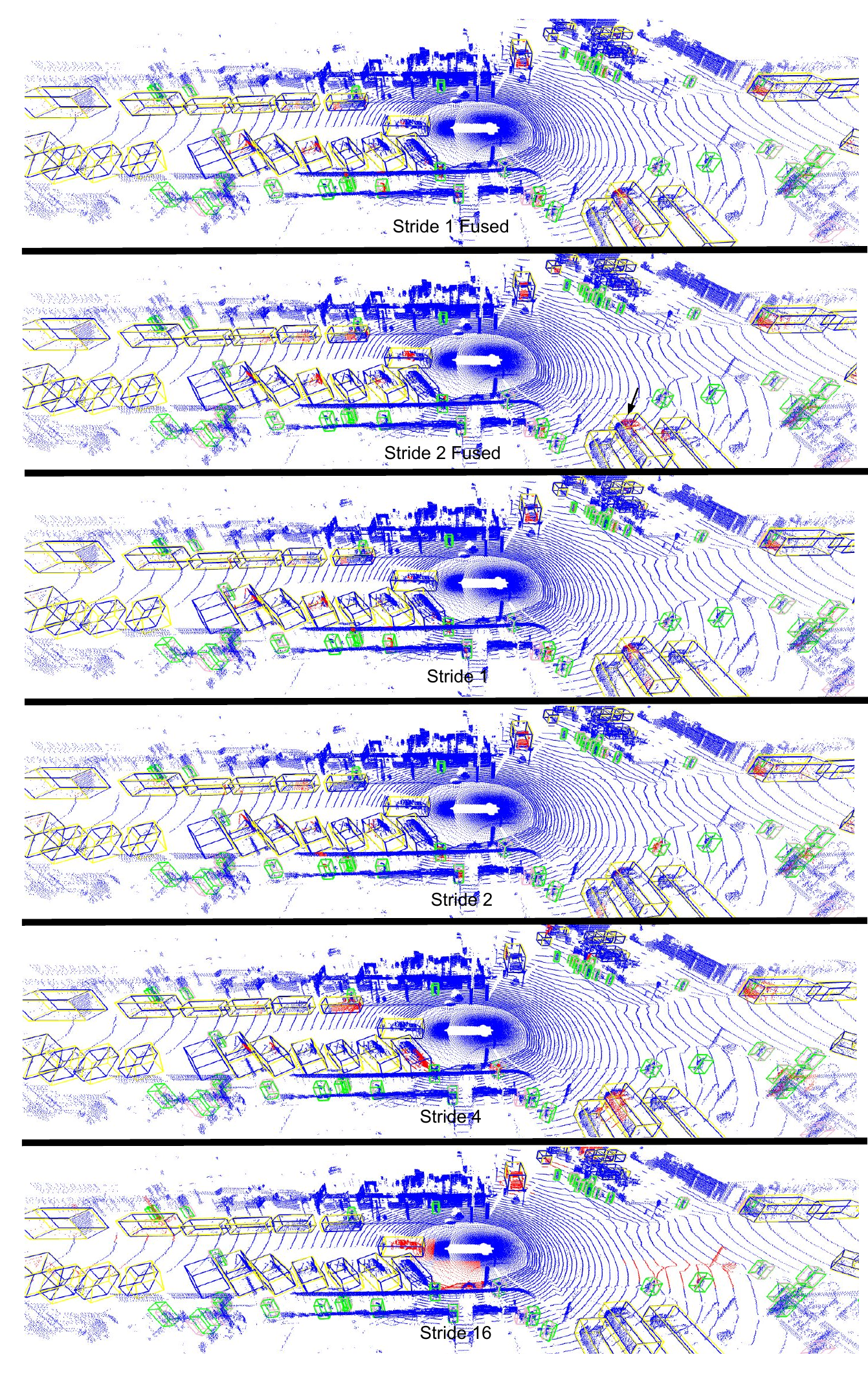}
   \caption{Attention scores and model prediction visualization. Blue box: ground truth vehicle. Yellow box: vehicle prediction. Green box: ground truth pedestrian. Purple box: pedestrian detection. Points are colored with the bwr colormap (0: blue, 1: red), where red points mean attention scores close to 1. As red points are  distributed differently in each subfigure, it is clear that different layers are attending to different locations.}
\label{fig:attention_score}
\end{figure}

\autoref{fig:attention_score} visualizes ground truth boxes, detected boxes, and attention scores for layers selected from different scales for the 15th frame in scene \\
8907419590259234067\_1960\_000\_1980\_000 selected from the Waymo Open Dataset validation set. The selected layers are the stride 1, 2 afters multi-scale feature fusion, and stride 1, 2, 4, 16 from the main backbone. We use all foreground points as the query points. The predicted boxes almost overlap perfectly with the ground truth boxes. The attention score pattern shown in these subplots indicates that different information is captured in different layers and scales. Interestingly, we have found that most of the attention scores are either 0 or 1 for foreground query points. We hope that these findings can inspire more research in the future.

\section{Future Work: More Tasks}
\begin{figure}[h!] 
\begin{center}
   \includegraphics[width=0.95\linewidth]{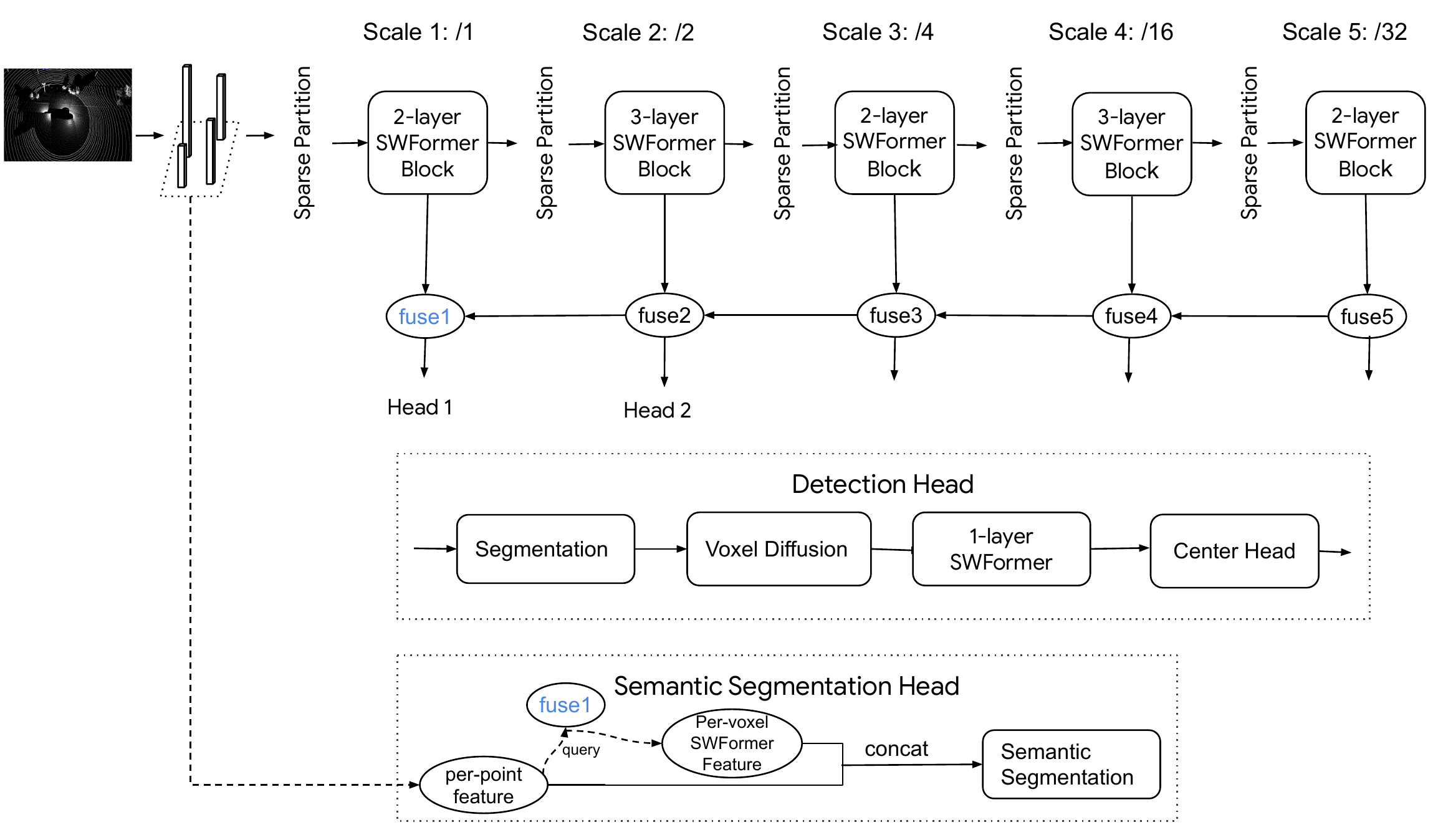} 
\end{center}
   \caption{Overview of the updated neural architecture for joint 3D detection and semantic segmentation. On top of \autoref{fig:arch}, it adds an extra segmentation head for the additional segmentation task.}
\label{fig:sem_seg_overview}
\end{figure}

\begin{figure}[h!] 
\begin{center}
   \includegraphics[width=0.8\linewidth]{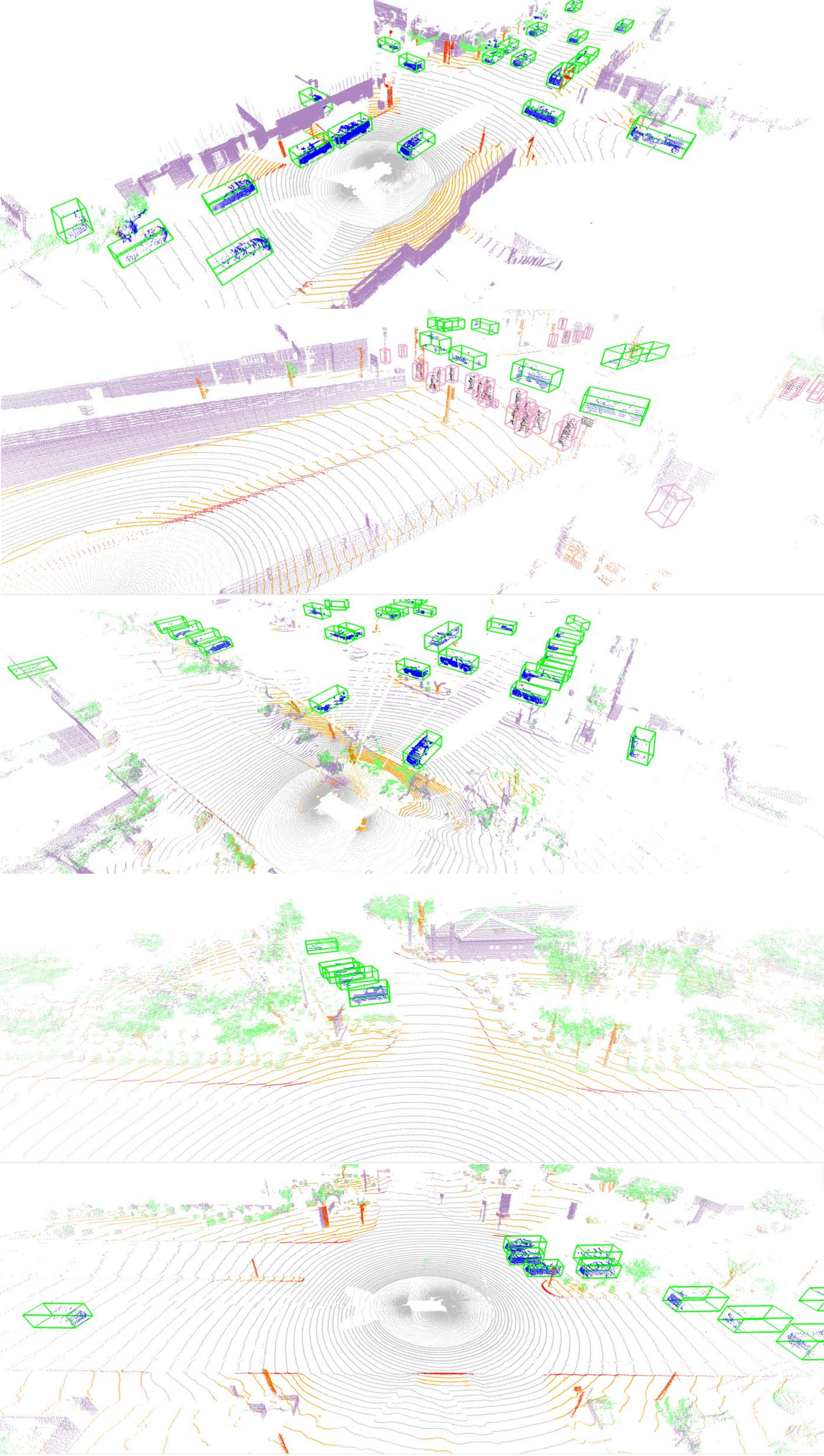}
\end{center}
   \caption{Joint detection and semantic segmentation qualitative results. Green boxes: vehicle. Lavender boxes: pedestrian. Lavender points: building. Grey points: road. Orange points: sidewalk. Blue points: vehicle. Black points: pedestrian. Red points: pole/sign/tree trunk. Green points: vegetation.}
\label{fig:sem_seg_result}
\end{figure}

Waymo Open Dataset~\cite{sun2020scalability} has recently added semantic segmentation labels for about $14\%$ of the frames per scene for all of the 1150 scenes. We have extended the SWFormer detection network to perform joint semantic segmentation and detection. Figure~\ref{fig:sem_seg_overview} illustrates the joint detection and semantic segmentation network architecture. We concatenate the per-point feature from the voxel embedding net before per-voxel max pooling and its corresponding voxel feature from a selected scale after multi-scale feature fusion to predict the per-point semantic segmentation logits. Without much tuning, we have obtained reasonable semantic segmentation results as shown in \autoref{tab:sem_seg} and \autoref{fig:sem_seg_result}. We plan to further improve this model and extend it to more autonomous driving related tasks.

\begin{table}[h]
\caption{Joint detection and semantic segmentation results on Waymo Open Dataset validation set and test set. }
\centering
\begin{tabular}{l|cc}
 Class Name & Validation IOU & Test IOU \\
\toprule
Bicycle & 36.76 & 38.15 \\
Bicyclist & 51.43 & 51.77 \\
Building & 75.18 & 65.75 \\
Bus & 65.45 & 39.50 \\
Car & 75.05 & 72.29 \\
Construction Cone & 48.34 & 21.37 \\
Curb & 55.54 & 48.46 \\
Lane Marker & 43.97 & 30.73 \\
Motorcycle & 56.68 & 58.37 \\
Motorcyclist & 1.48 & 0.57 \\
Other Ground & 34.34 & 37.52 \\
Other Vehicle & 23.95 & 25.43 \\
Pedestrian & 60.87 & 61.08 \\
Pole & 55.50 & 51.65 \\
Road & 78.46 & 68.06 \\
Sidewalk & 59.67 & 59.77 \\
Sign & 53.70 & 43.60 \\
Traffic Light & 22.74 & 22.30 \\
Tree Trunk & 54.74 & 50.64 \\
Truck & 48.73 & 55.86 \\
Vegetation & 79.78 & 68.08 \\
Walkable & 65.87 & 59.08 \\
\midrule
mIOU & 52.19 & 46.82 \\
\end{tabular}
\label{tab:sem_seg}
\end{table}

\end{document}